\documentclass[10pt,twocolumn]{article}
\usepackage[letterpaper,margin=0.65in,columnsep=0.28in]{geometry}
\usepackage[T1]{fontenc}
\usepackage[utf8]{inputenc}
\usepackage{lmodern,microtype}
\usepackage{amsmath,amssymb,amsthm,mathtools}
\usepackage{graphicx,booktabs,array,tabularx,longtable}
\usepackage{enumitem,stfloats}
\usepackage[numbers,sort&compress]{natbib}
\usepackage[hidelinks]{hyperref}
\hypersetup{pdftitle={The geometry of AI validation: From structural blindness to reusable audits},pdfauthor={Ricardo Fitas},pdfsubject={AI validation, statistical identification, and audit complexity}}
\usepackage[font=small,labelfont=bf]{caption}
\usepackage{fancyhdr}
\setlist{nosep,leftmargin=*}
\newtheorem{theorem}{Theorem}

\newcommand{\E}{\mathbb E}
\newcommand{\Prb}{\mathbb P}
\newcommand{\R}{\mathcal R}
\newcommand{\TV}{\operatorname{TV}}
\newcommand{\esssup}{\operatorname*{ess\,sup}}

\newcommand{\Var}{\operatorname{Var}}
\newcommand{\KL}{\operatorname{KL}}
\newcommand{\sgn}{\operatorname{sgn}}
\newcommand{\one}{\mathbf 1}

\begin{document}
\twocolumn[{\begin{@twocolumnfalse}
\begin{center}
{\fontsize{21}{24}\selectfont\bfseries The geometry of AI validation:\\From structural blindness to reusable audits\par}
\vspace{0.7em}
{\large Ricardo Fitas\par}
\vspace{0.25em}
Technical University of Darmstadt, 64289 Darmstadt, Germany\\
\href{mailto:rfitas99@gmail.com}{rfitas99@gmail.com}
\end{center}
\vspace{0.6em}
\noindent\textbf{Abstract.}
AI systems increasingly search among candidate answers and deploy the highest-scoring one. Increasing search changes which errors matter, so a precise evaluation at one computation budget can leave another budget unresolved. We connect this information gap to the cost of closing it. For independent best-of-$n$ search, aggregate reliability measurements identify deployment only through the directions they observe; we derive an exact ambiguity frontier when only small search widths are audited. Retaining candidate ranks and truth labels enables a constructive alternative: one audit can estimate reliability across all widths up to $N$. With known score percentiles, the minimax worst-coordinate mean squared error scales as $(1+\log N)/T+N/M$, capped at a constant, for expected budgets of $T$ truth labels and $M$ candidate observations. Matching lower bounds allow adaptive label acquisition, establishing that the distinct label and candidate costs are intrinsic to this experiment. An explicit design attains this order; a complementary record-based procedure supplies simultaneous guarantees without a known score distribution. Retrospective mathematical-reasoning and code-generation analyses show why search-dependent validation matters. In held-out CodeRM pools, a shared audit reduces the 95th-percentile maximum error across 100 widths by 58\% and 40\% relative to uniform labeling. These results turn structural ambiguity into a quantitative prescription for reusable validation.
\par\medskip
\noindent\textbf{Significance.}
A system that searches harder may become more accurate on average while becoming less reliable on particular problems. Evaluating a small search does not necessarily reveal what happens after a larger one. We show how this missing information can be acquired and reused across search budgets. The key is to retain where verified answers occur in the score distribution. For a fixed mean squared error in an independent-search model, the required truth-label budget grows logarithmically with search breadth, while the candidate budget can grow linearly, even for adaptive audits. This separation identifies when expensive verification can be saved and when access to candidates remains the bottleneck.
\vspace{0.8em}
\end{@twocolumnfalse}}]

\section{Introduction}
An AI search system generates candidates, ranks them, and deploys the highest-scoring output. Its reliability is the truth of that selected output. A benchmark average and a deployment search can therefore place very different weight on the same errors.

This distinction is visible even when search works well overall. Repeated sampling can improve performance \citep{brown2024}, while optimizing an imperfect reward can produce poor outcomes despite low average error \citep{fluri2025}. Verifier ROC geometry characterizes iid selection and limits extrapolation from small to large search budgets \citep{dorner2026}; coverage also shapes the performance of inference-time alignment \citep{huang2025}. These results motivate an operational question: what evidence should an evaluator collect to support several possible deployment budgets?

We connect identification to acquisition. An aggregate audit observes an average of an unknown reliability surface; repeating it improves precision in the same direction. Candidate ranks and truth evaluations can reveal unresolved directions, which concentrate on progressively rarer score tails as search grows.

We calculate the ambiguity remaining after every small-width mean is known, then derive the candidate and truth budgets needed to estimate the full search family. Lower bounds allow adaptive acquisition; an explicit sampling rule attains the minimax order. Simultaneous bands protect a width chosen after evaluation. Two empirical domains connect these results to observable behavior.

Optimal recovery and polynomial approximation provide the identification tools \citep{donoho1994,bojanov1986,newman1976}. Shared-policy evaluation, information-radius designs, and sequential change of measure provide the acquisition tools \citep{chen2025,nakiboglu2019,kaufmann2016,russo2025}. Our focus is their quantitative connection for the full best-of-$n$ family: an explicit ambiguity frontier, a minimax law with two separately counted resources, and realizable audits.

\section{Results}
\subsection{Search changes the reliability target in two domains}
We first evaluated two public candidate collections with objective correctness labels. The mathematical-reasoning analysis contains 127 GSM8K problems per model and 2.54 million generated solutions in total \citep{brown2024}. Answer frequencies in a separate reference split provide consensus scores. The code analysis contains 164 HumanEval+ tasks per model and 32,800 programs; generated unit tests provide scores and separate HumanEval+ tests provide truth \citep{ma2025}. Both analyses evaluate the exact with-replacement search functional of each fixed task pool (Materials and methods).

\begin{figure*}[t]
\centering\includegraphics[page=1,width=0.94\textwidth]{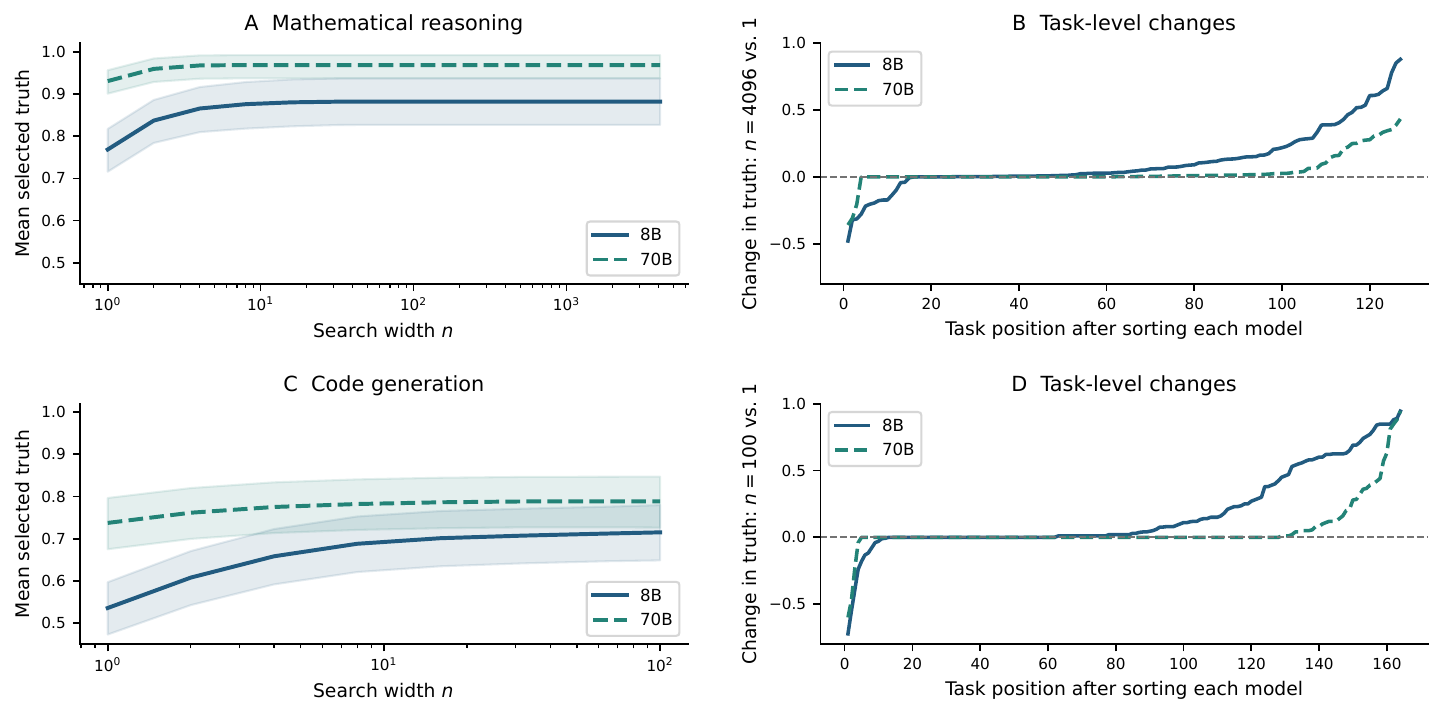}
\caption{\textbf{Search improves the mean while changing which problems fail.} (A,C) Mean selected truth in mathematical reasoning and code generation; shading denotes 95\% problem-bootstrap intervals. (B,D) Sorted task-level changes between the smallest and largest displayed widths. Each model is sorted separately; positions do not identify matched tasks. Dashed horizontal lines mark zero change. The fixed pools contain 127 mathematical problems and 164 code tasks per model. All curves use exact with-replacement selection with randomized score ties.}
\label{fig:domains}
\end{figure*}

Mean correctness increased with search in both domains (Fig.~\ref{fig:domains}). From widths 1 to 4,096, GSM8K means rose from 0.768 to 0.882 for 8B and from 0.931 to 0.968 for 70B. Nevertheless, 14 and three problems lost more than one percentage point, with worst losses of 48 and 36 points. The harm counts persisted across ten reference/deployment splits. In code, widths 1 to 100 increased means from 0.536 to 0.715 and from 0.737 to 0.788, while ten and four tasks lost more than one point; the largest losses were 72 and 60 points. Thus favorable aggregate scaling coexists with substantial changes in individual reliability targets.

\subsection{Aggregate audits leave an exact information gap}
Let $U$ be a candidate's randomized score percentile and $Y\in\{0,1\}$ its truth. Under iid generation, stable scoring, and an unchanged rank--truth law,
\begin{align}
 f(u)&=\Prb(Y=1\mid U=u),\quad k_n(u)=nu^{n-1},\notag\\
 \theta_n&=\int_0^1 k_n(u)f(u)\,du.
 \label{eq:target}
\end{align}
Here $U$ is uniform and ties are resolved with independent continuous keys. No shape or calibration assumption is imposed on $f$. A validation rule with kernel $v$ observes $\int vf$; deployment with kernel $k$ requires $\int kf$.

\begin{theorem}[Identification by audit geometry]
If $V$ is the span of finitely many audited kernels, the largest target separation between bounded reliability laws agreeing on every audit is
\begin{equation}
 D(k,V)=\inf_{v\in V}\|k-v\|_1.
 \label{eq:span}
\end{equation}
For exact reliability means at every width $1,\ldots,m<N$, this separation is
\begin{equation}
 B_{m,N}=1+2\sum_{r=1}^m(-1)^r
 \cos^{2N}\!\left(\frac{r\pi}{2(m+1)}\right).
 \label{eq:blind}
\end{equation}
The central audit vector $(1/2,\ldots,1/2)$ attains the full interval $[(1-B_{m,N})/2,(1+B_{m,N})/2]$.
\label{th:identification}
\end{theorem}
The proof constructs bounded complementary reliability laws using a Chebyshev-sign function (SI Section S1). The governing scale is $m^2/N$: when this ratio tends to one, ambiguity tends to approximately 0.83 (Fig.~\ref{fig:geometry}A). Even exact replication of the same small-width means cannot close that gap. Monotone Lipschitz reliability laws can also retain ambiguity; SI Section S10 gives constructive bounds. The statement concerns aggregate means; $m$ is not a label count.

The gap also appears for an empirical audit vector. Retaining only CodeRM means at widths 1--8 permits bounded step-function laws that agree on those means yet yield best-of-100 reliabilities spanning $[0.186,0.992]$ for 8B and $[0.305,0.996]$ for 70B. Their observed targets are 0.715 and 0.788. These intervals are feasible witnesses within a discretized subclass, not claims that such extreme worlds are typical. SI Section S9 reports the construction and sensitivity checks.

\subsection{The missing information has a minimax acquisition cost}
Candidate-level observations change the available experiment. Suppose population score percentiles are known. An audit can inspect iid candidates, request selected truth labels, and adapt both acquisition and stopping to the observed history. Let $M$ and $T$ bound the expected numbers of inspected candidates and requested labels, uniformly over $f$, with $M\ge T\ge1$. A candidate contributes at most one independently revealed truth label.

Define the best attainable worst-coordinate squared error by
\begin{equation}
 \mathfrak R_N(M,T)=\inf_{\mathcal A}\sup_{0\le f\le1}
        \max_{n\le N}\E_f(\widehat\theta_n-\theta_n)^2,
 \label{eq:minimax}
\end{equation}
where the infimum includes all such adaptive audits and all estimators.

\begin{figure*}[t]
\centering\includegraphics[page=2,width=\textwidth]{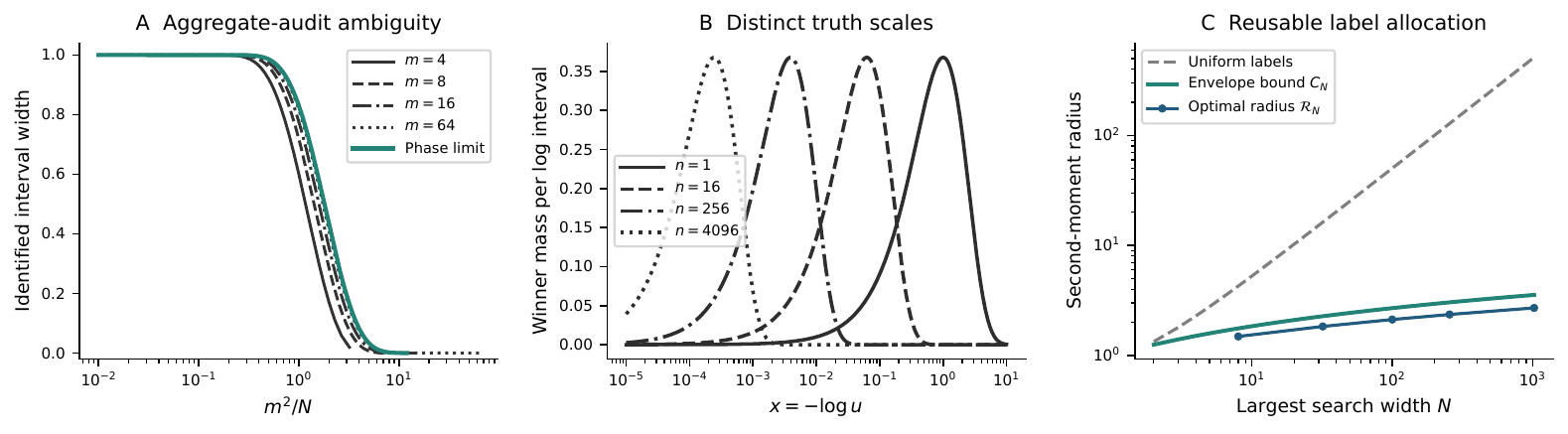}
\caption{\textbf{The geometry connects missing information to the cost of acquiring it.} (A) Exact aggregate-audit ambiguity approaches a common curve in $m^2/N$. (B) In the coordinate $x=-\log u$, winner densities are exponential; their mass per logarithmic interval, $xp_n(x)$, shifts across scales. Geometrically spaced widths can encode separate truth questions. (C) Uniform labeling has a variance radius growing linearly with $N$, while the envelope and optimal designs grow logarithmically. Numerical radius values use finite lower and upper brackets, whose gaps are smaller than the symbols.}
\label{fig:geometry}
\end{figure*}

\begin{theorem}[Two-resource minimax law]
For integer $N\ge2$ and $T\ge1$, and real $M\ge T$,
\begin{equation}
 \mathfrak R_N(M,T)
 =\Theta\!\left(\min\left\{1,\frac{1+\log N}{T}+\frac{N}{M}\right\}\right),
 \label{eq:fullfrontier}
\end{equation}
with universal constants. The upper bound is achieved by a label-independent sampling rule. The lower bound allows adaptive label selection and stopping.
\label{th:adaptive}
\end{theorem}
This result separates two bottlenecks. Broadening search creates logarithmically many distinguishable tail scales, each of which can contain independent truth information (Fig.~\ref{fig:geometry}B). Even an audit that actively chooses the next rank cannot learn all those scales from a constant label budget. Separately, an error confined to the top $1/N$ fraction of candidates has order-one influence after width-$N$ search. Discovering its truth requires candidate access that grows linearly with $N$. The latter lower bound holds even when every observed candidate is labeled for free and its score percentile is known.

Consequently, for a fixed small mean squared error $\varepsilon^2$, the optimal expected cost has order
\begin{equation}
 \frac{c_Y(1+\log N)+c_XN}{\varepsilon^2},
 \label{eq:economics}
\end{equation}
where $c_Y$ is the additional cost of truth and $c_X$ the cost of obtaining and scoring a candidate. SI Section S7 gives explicit finite constants and the adaptive lower-bound proofs. Equation~\ref{eq:fullfrontier} concerns the largest coordinatewise expected error; simultaneous confidence is treated separately below.

\subsection{One score-based design serves every search width}
The constructive audit samples ranks from a density $q$ and uses the same labels for every target:
\begin{equation}
 \widehat\theta_n=\frac12+\frac1T\sum_{i=1}^T
       \frac{k_n(U_i)}{q(U_i)}(Y_i-\tfrac12).
 \label{eq:estimator}
\end{equation}
Its worst squared error is exactly $\max_n V_n(q)/(4T)$, where $V_n(q)=\int k_n^2/q$. An explicit choice is the normalized envelope
\begin{align}
 q_N^{\mathrm{env}}(u)&=\frac{\max_{n\le N}nu^{n-1}}{C_N},\notag\\
 C_N&=1+\sum_{j=1}^{N-1}\frac{j^j}{(j+1)^{j+1}}
       =e^{-1}\log N+O(1).
 \label{eq:envelope}
\end{align}
It bounds every importance weight by $C_N$: the same audit covers an entire range of search widths. The normalized-envelope principle follows the classical minimax likelihood-ratio construction \citep{shtarkov1987}; here the full integer search family has an explicit normalizer.

Variance optimization gives a smaller radius,
$\R_N:=\inf_q\max_nV_n(q)=\tfrac14\log N[1+o(1)]$.
Its finite values can be found through a concave dual \citep{nakiboglu2019,chen2025}. At $N=100$, $\R_N\approx2.1173$ and $C_N\approx2.6873$, compared with $N^2/(2N-1)\approx50.25$ for uniform labels. These radii describe unrestricted percentile laws; score ties in finite pools can reduce them further. Proofs, endpoint corrections, and density-capped constructions are in SI Sections S2--S4.

To realize the design with ordinary candidates, accept a scored candidate with probability $q(U)/\kappa$ for a density cap $q\le\kappa$. Each accepted label requires $\kappa$ candidate draws in expectation. A mixture of the uniform and envelope designs accommodates $\kappa=M/T$ and proves the upper bound in Theorem~\ref{th:adaptive}.

\subsection{Reusable audits improve complete-curve estimation}
We tested five score-based designs on a shared 82-task held-out CodeRM pool per model: uniform labeling, a top-5\% mixture, a winner-100 mixture, the envelope, and variance minimization. The two focused mixtures retain 20\% uniform coverage, so all widths remain supported. The entire score pool determines each proposal; its evaluation labels do not enter optimization.

\begin{figure*}[t]
\centering\includegraphics[page=3,width=\textwidth]{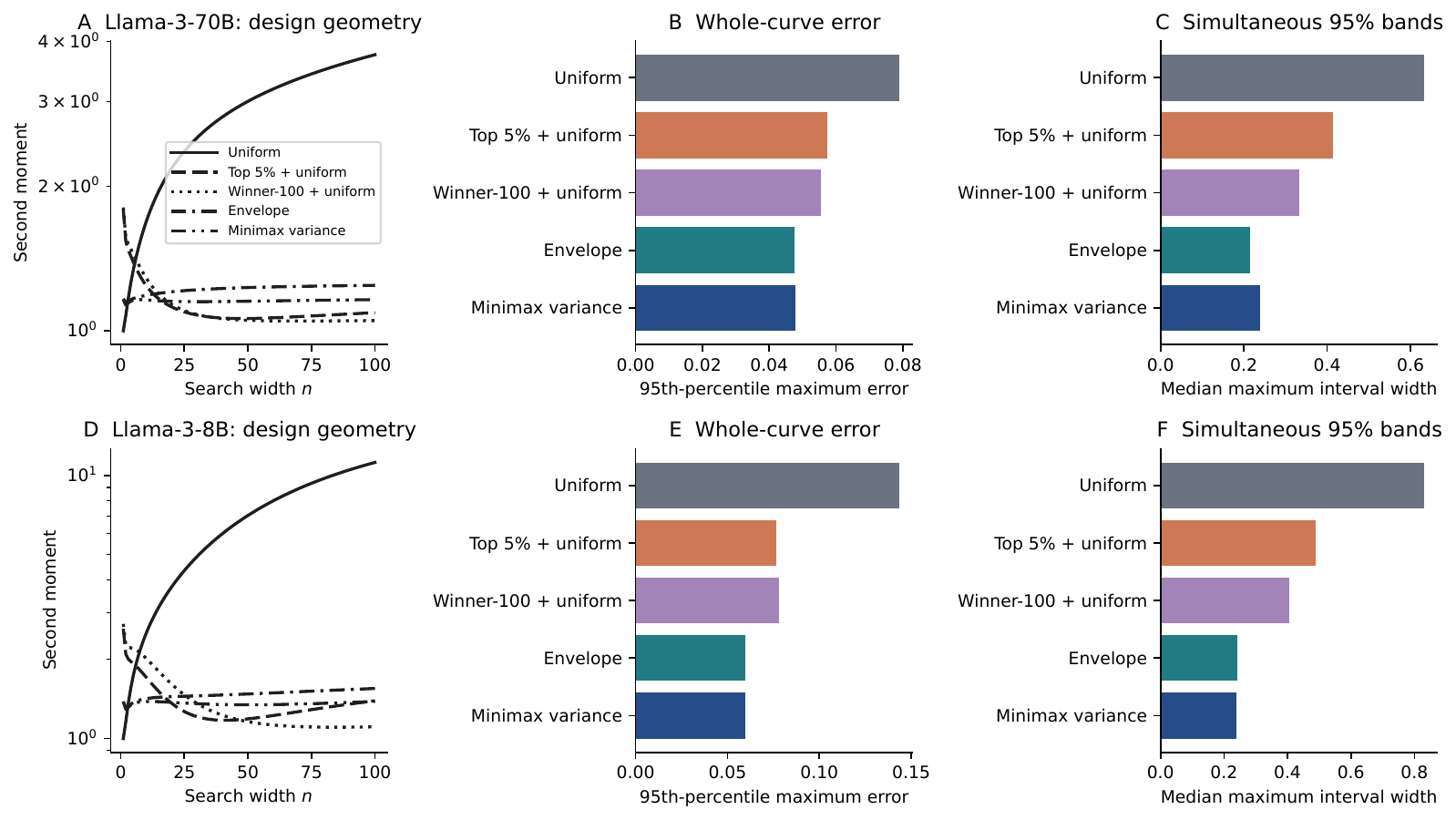}
\caption{\textbf{A shared audit improves estimation across the complete search curve.} Left: second moments across widths, distinguished by line style. Middle: the 95th percentile of maximum absolute error across all 100 widths at 500 label draws. Right: the median maximum width of simultaneous 95\% confidence bands, clipped to $[0,1]$. Results use 2,000 retrospective acquisition replays per design in the fixed 82-task held-out pools. The left panels depend only on scores; error and coverage use independently specified benchmark truth.}
\label{fig:empirical}
\end{figure*}

At 500 label draws and 2,000 acquisition replays, the envelope reduced the 95th-percentile maximum absolute error over widths 1--100 from 0.143 to 0.060 for 8B and from 0.079 to 0.048 for 70B (Fig.~\ref{fig:empirical}). The two focused mixtures gave errors of 0.077--0.078 and 0.056--0.057, respectively. Thus the advantage persists against score-targeted comparators. Envelope and variance-optimal designs had similar observed errors; on 70B, the envelope produced narrower simultaneous bands because its weight control also matters.

A complementary fixed-target experiment selected a score-tail fraction on 82 discovery tasks and froze it before evaluation on the held-out tasks. At the same 500-label budget, best-of-100 error decreased from 0.180 to 0.049 for 8B and from 0.076 to 0.035 for 70B. That experiment estimates one width and uses a different estimator; its numbers are not directly comparable with the complete-curve metric. Together, the experiments distinguish a focused audit for a fixed target from a shared audit for a selectable family (SI Sections S8--S9).

\subsection{Simultaneous guarantees support choices after the audit}
The curve can guide the choice of search width only if its uncertainty accounts for that choice. Bernstein bands for Eq.~\ref{eq:estimator}, combined across the target family, provide finite-sample simultaneous coverage. In the CodeRM replays, all methods covered the whole curve in every repetition; their analytic guarantee is at least 95\%, and the intervals are conservative. The observed improvement is lower error and narrower bands at the same label budget.

When population percentiles are unknown, another construction labels each new score record along an iid search path. That record's truth is reused until a new winner appears. A path of $N$ candidates requires $H_N=\sum_{n=1}^N1/n$ record labels on average \citep{nevzorov2001}. Independent paths yield valid confidence bands. Since
\begin{equation}
 |\theta_m-\theta_n|\le e^{-1}|\log(m/n)|,
 \label{eq:modulus}
\end{equation}
a logarithmic grid suffices to cover every integer width. For simultaneous error at most $\varepsilon$ with probability $1-\alpha$, this construction uses expected resources
\begin{align}
 T&=O\!\left(\frac{1+\log N}{\varepsilon^2}
               \log\frac{2+\log N/\varepsilon}{\alpha}\right),\notag\\
 M&=O\!\left(\frac{N}{\varepsilon^2}
               \log\frac{2+\log N/\varepsilon}{\alpha}\right).
 \label{eq:confidencecost}
\end{align}
These guarantees protect any width selected from the audited family. The record replay matched the harmonic label count, with lower precision than the fully scored-pool designs; it demonstrates the tradeoff associated with dispensing with population percentiles (SI Section S8).

\section{Discussion}
The central distinction is between repeating an observation and acquiring a missing direction of evidence. In the aggregate experiment, exact low-width means can coexist with sharply different deployment reliabilities. In the candidate experiment, recording rank and truth supplies reusable evidence across search scales. Theorem~\ref{th:adaptive} shows that the resulting logarithmic label cost is intrinsic even when acquisition adapts to the answers already obtained.

This suggests a concrete validation strategy: specify the generator, scorer, task population, and selectable computation budgets; retain candidate ranks alongside independently assessed truth; and allocate labels across the relevant tail scales. Uncertainty should cover the choice subsequently made. Budget candidate access and truth separately: expensive truth need not be requested for every generated answer.

The empirical advantage is conditional on existing CodeRM pools; prospective monetary savings and transfer to new generators remain unmeasured. The minimax upper bound assumes known percentiles. The record construction removes that requirement at an additional confidence factor. Correlated generation, changing scorers, and unreliable truth assessments require explicit models beyond the iid experiment.

Several questions follow directly. The sharp constants for fully adaptive audits remain unknown, as does the necessity of the additional logarithmic factor in simultaneous confidence. A prospective comparison should measure candidate and truth costs at a common error and coverage target. Extending the geometry to history-dependent generators requires representing how each policy changes the selected-output distribution.

\section{Materials and methods}
\paragraph{Selection model and inferential units.}
Candidates are iid conditional on a task. Each draw has a score and binary truth, with independent continuous tie keys. For task $t$, winner truth is $\theta_{n,t}=\int k_nf_t$; equal task averaging gives the macro target. All empirical search curves refer to draws with replacement from fixed task pools. Problems provide the independent units for the bootstrap summaries; candidate replays quantify acquisition randomness conditional on the held-out pools.

\paragraph{Mathematical reasoning.}
The pinned Monkey Business release supplies 10,000 GSM8K solutions for each of 127 problems per model. A seeded split assigns 2,000 solutions to reference scoring and 8,000 to the deployment pool. Normalized answer frequency in the reference set supplies the consensus score; exact final-answer correctness supplies truth. Search curves use exact score-tie integration, 20,000 problem-bootstrap replicates, and ten reference/deployment splits for robustness.

\paragraph{Code generation and acquisition.}
CodeRM scores are fractions of generated unit tests passed; truth uses HumanEval+ \texttt{plus\_status}. The shared split assigns 82 tasks each to discovery and held-out evaluation. Complete-curve proposals use held-out scores without evaluation labels. The primary metric is the 95th percentile of maximum error across widths 1--100, with 2,000 replays at 500 labels. Distinct program counts are reported separately. Fixed-target acquisition uses 5,000 replays and discovery selection among tail fractions 5\%, 10\%, and 20\%.

\paragraph{Verification and reproducibility.}
The SI supplies proofs, numerical brackets, confidence formulas, and comparator results. The code package supplies processed inputs, raw-data reconstruction scripts, source identifiers, file hashes, and random seeds.

\section{Conclusions}
Validation should track the distribution of decisions that search will produce. Aggregate evidence has an exact geometric boundary; candidate-level evidence can cross it. For iid search with known percentiles, the optimal candidate and truth budgets have different scaling laws, and adaptive acquisition cannot remove either bottleneck in the worst case. Explicit shared audits and record-based certificates make this distinction operational, allowing verified evidence to support a family of search budgets.

\paragraph{Data and code availability.}
Data, results, source identifiers, and reproduction code accompany this article. Project repository: \href{https://github.com/rfitas-lab/geometry-of-ai-validation}{rfitas-lab/geometry-of-ai-validation}.

\clearpage\onecolumn

\begin{center}
{\LARGE\bfseries Supporting Information\par}
\vspace{0.6em}
{\large The geometry of AI validation: From structural blindness to reusable audits\par}
\end{center}
\renewcommand{\theequation}{S\arabic{equation}}\setcounter{equation}{0}
\renewcommand{\thefigure}{S\arabic{figure}}\setcounter{figure}{0}
\renewcommand{\thetable}{S\arabic{table}}\setcounter{table}{0}

\section*{S1. Structural identification and the exact prefix frontier}
\paragraph{The span identity.}
Let $V$ be a finite-dimensional subspace of $L^1(\mu)$ and let reliability laws lie in $\mathcal F=\{f\in L^\infty:0\le f\le1\}$. The difference set $\mathcal F-\mathcal F$ is exactly the unit ball of $L^\infty$: write any $h$ with $\|h\|_\infty\le1$ as $f-g$ with $f=(1+h)/2$, $g=(1-h)/2$. Consequently,
\begin{equation}
 \sup_{\substack{f,g\in\mathcal F\\\langle v,f-g\rangle=0\;(v\in V)}}
 |\langle k,f-g\rangle|
 =\sup_{\substack{h\in V^\perp\\\|h\|_\infty\le1}}|\langle k,h\rangle|
 =\inf_{v\in V}\|k-v\|_1.
\end{equation}
The last equality is quotient/annihilator duality; Hahn--Banach supplies an attaining functional because $V$ is closed. This is the optimal-recovery geometry associated with the bounded reliability class \citep{donoho1994}. If all kernels are normalized, the evidence consisting entirely of $1/2$ has endpoints $(1\pm D(k,V))/2$. Scaling an extremizer fills the interval. At other evidence vectors the feasible interval can be smaller.

\paragraph{Winner kernels and canonical approximation.}
The randomized probability-integral transform makes a candidate's rank $U$ uniform. The maximum among $n$ iid ranks has CDF $u^n$, hence density $k_n(u)=nu^{n-1}$. Therefore the means at widths $1,\ldots,m$ span $\mathcal P_{m-1}$.
Put $\sigma_m(u)=\sgn U_m(2u-1)$, where $U_m$ is the Chebyshev polynomial of the second kind. For $q\in\mathcal P_{m-1}$, substitute $2u-1=\cos\theta$ into $\int q\sigma_m$. The factor $q((1+\cos\theta)/2)\sin\theta$ is a linear combination of $\sin(\ell\theta)$, $1\le\ell\le m$. The $L^2$ sine series of $\sgn\sin((m+1)\theta)$ has only frequencies $(2r+1)(m+1)$. Orthogonality gives
\begin{equation}
 \int_0^1q(u)\sigma_m(u)\,du=0.
 \label{si:annihilation}
\end{equation}
For $N>m$, interpolate $k_N$ at $u_j=(1+\cos(j\pi/(m+1)))/2$, $j=1,\ldots,m$, by $p\in\mathcal P_{m-1}$. Its divided-difference remainder has sign $\sigma_m$ almost everywhere because $k_N^{(m)}>0$ on $(0,1]$. Thus, for every $q\in\mathcal P_{m-1}$,
\begin{equation}
 \|k_N-q\|_1\ge\int\sigma_m(k_N-q)
 =\int\sigma_m k_N=\|k_N-p\|_1.
\end{equation}
This is the canonical-point construction in classical $L^1$ approximation \citep{bojanov1986}. Integrating over its alternating sign intervals gives
\begin{equation}
 \inf_{p\in\mathcal P_{m-1}}\|k_N-p\|_1
 =1+2\sum_{r=1}^m(-1)^r
 \cos^{2N}\!\left(\frac{r\pi}{2(m+1)}\right)=B_{m,N}.
\end{equation}
The laws $f_\pm=(1\pm\sigma_m)/2$ have all audited means equal to $1/2$ and deployment means $(1\pm B_{m,N})/2$. Convex mixtures of their conditional Bernoulli laws fill the identified interval. For $N\le m$, the target already belongs to the audited span and the diameter is zero.

\paragraph{Audit schedules and the phase coordinate.}
Every randomized mixture of reliability-mean queries at widths at most $m$ lies in $\mathcal P_{m-1}$. The complete prefix spans that space. Even sequential choice of such exact mean queries cannot distinguish $f_+$ and $f_-$, which return $1/2$ to every allowed query. Candidate ranks, candidate truth observations, and score-dependent stopping can add information beyond this query class.

When $m,N\to\infty$ with $m^2/N\to\tau\in(0,\infty)$, termwise convergence gives
\begin{equation}
 B_{m,N}\longrightarrow B(\tau)
 =1+2\sum_{r=1}^{\infty}(-1)^r e^{-\pi^2r^2/(4\tau)}
 =4\sqrt{\tau/\pi}\sum_{j=0}^{\infty}e^{-\tau(2j+1)^2}.
 \label{si:phase}
\end{equation}
For a rigorous exchange of sum and limit, use $\cos x\le e^{-x^2/2}$ to dominate the $r$th summand eventually by $\exp[-\pi^2r^2/(8\tau)]$. The last expression follows from the Jacobi transformation \citep{whittaker1927}. It gives $B(1)\approx0.8305$ and $B(\tau)\sim4\sqrt{\tau/\pi}e^{-\tau}$ as $\tau\to\infty$. Thus a small limiting diameter requires a phase coordinate of order $\log(1/\varepsilon)$. This inversion takes the fixed-phase limit before sending $\varepsilon$ to zero; it is not a uniform remainder statement for arbitrary joint limits. The degree-squared approximation scale has classical antecedents \citep{newman1976}.

\section*{S2. Exact envelope design}
\paragraph{Envelope identity.}
Suppose $W(q)\le c$. For almost every $u$, $k_n(u)\le cq(u)$ for every $n$, hence $g_N(u)\le cq(u)$. Integration gives $C_N:=\int g_N\le c$. The density $q=g_N/C_N$ attains equality. This is the normalized maximum-likelihood/minimax-ratio construction \citep{shtarkov1987}; its normalizer for the integer search family is explicit.

\paragraph{The maximizing width.}
For $u\in(0,1)$,
\begin{equation}
 \frac{k_{n+1}(u)}{k_n(u)}=\frac{n+1}{n}u.
\end{equation}
As $n$ increases, the ratio decreases. Therefore $k_n$ is maximal on
$[(n-1)/n,n/(n+1)]$, with the initial interval beginning at zero and the final interval for $n=N$ ending at one. Write $a_j=j/(j+1)$. Integrating the maximizing density on each interval yields
\begin{align}
 C_N&=1+\sum_{j=1}^{N-1}(a_j^j-a_j^{j+1})\\
    &=1+\sum_{j=1}^{N-1}\frac{j^j}{(j+1)^{j+1}}.
\end{align}
Since the $j$th summand is $e^{-1}/j+O(j^{-2})$, the stated logarithmic asymptote follows. For the continuous family $1\le s\le N$, the maximizing rate in $x=-\log u$ is $1$, $1/x$, or $N$ on the three intervals cut by $1$ and $1/N$. Its envelope normalizer is exactly $1+e^{-1}\log N$, providing a useful finite upper check on the integer formula.

\paragraph{Numerical checks.}
Closed sums and independent piecewise adaptive quadrature were compared at $N\in\{1,2,3,8,16,64,100,256\}$. Together with total-variation and variance-identity checks, the maximum discrepancy was $2.56\times10^{-15}$. These are arithmetic diagnostics, not substitutes for the proofs.

\section*{S3. Variance radius: exact risk, dual, and sharp asymptotics}
\paragraph{Exact risk.}
For a single observation, let $X_n=k_n(U)(Y-1/2)/q(U)$. Since $Y$ is binary,
$(Y-1/2)^2=1/4$, regardless of its conditional success probability. Hence
\begin{equation}
 \E X_n=\theta_n-\frac12,\qquad
 \Var(X_n)=\frac14V_n(q)-(\theta_n-\tfrac12)^2.
 \label{si:variance}
\end{equation}
Independence across audit draws gives the upper bound in the main-text risk identity. The single surface $f\equiv1/2$ attains the bound for every target simultaneously. Notice the order of the loss: this is $\sup_f\max_n\E(\widehat\theta_n-\theta_n)^2$, not $\sup_f\E\max_n(\widehat\theta_n-\theta_n)^2$.

\paragraph{Dual design.}
For $\lambda\in\Delta_N$, write $s_\lambda=\sum_n\lambda_n k_n^2$ and $Z_\lambda=\int\sqrt{s_\lambda}$. Cauchy--Schwarz implies
\begin{equation}
 \max_nV_n(q)\ge\sum_n\lambda_nV_n(q)
   =\int s_\lambda/q\ge Z_\lambda^2.
\end{equation}
The maximum of $Z_\lambda^2$ exists because the simplex is compact and the integral is continuous. At a maximizing $\lambda^*$, put $q^*=\sqrt{s_{\lambda^*}}/Z_{\lambda^*}$. The directional derivative toward the $n$th simplex vertex is
$V_n(q^*)-Z_{\lambda^*}^2$ and must be nonpositive. An infinite positive derivative would also contradict maximality. Thus all $V_n(q^*)\le Z_{\lambda^*}^2$, proving equality. Active coordinates have equality. This is a short specialization of the order-two R\'enyi capacity/radius dual \citep{nakiboglu2019}. In discrete problems the same proof replaces integrals by sums.

\paragraph{Transformation.}
Under $x=-\log u$, target and audit measures become
\begin{equation}
 p_n(x)=ne^{-nx},\qquad \widetilde q(x)=q(e^{-x})e^{-x}.
\end{equation}
Their likelihood ratios and second moments are invariant under the transformation. We establish both directions of the radius asymptote for the \emph{integer} family $n=1,\ldots,N$.

\paragraph{Integer-family lower bound.}
Use the discrete prior $\lambda_n=(nH_N)^{-1}$. Then
\begin{equation}
 \R_N\ge\frac1{H_N}\left(\int_0^\infty
            \sqrt{\sum_{n=1}^N ne^{-2nx}}\,dx\right)^2.
 \label{si:discreteLB}
\end{equation}
The infinite sum is $(2\sinh x)^{-2}$. Its finite truncation is the infinite sum multiplied by
\begin{equation}
 1-e^{-2Nx}\{1+N(1-e^{-2x})\}.
\end{equation}
Let $L=\log N$ and restrict the integral to $x\in[L/N,1/L]$. Uniformly on this interval the multiplier is $1-o(1)$, because its deficit is at most $(1+2L)e^{-2L}$, and $\sinh x=x[1+o(1)]$. Therefore the integral in Eq.~\ref{si:discreteLB} is at least
\begin{equation}
 \frac12(1-o(1))\int_{L/N}^{1/L}\frac{dx}{x}
 =\frac12(1-o(1))(\log N-2\log\log N).
\end{equation}
Since $H_N\sim\log N$, this proves $\liminf\R_N/\log N\ge1/4$.

\paragraph{Uniform upper bound, including endpoints.}
Choose $a=a_N\to\infty$ with $a_N=o(\log N)$; for example $a_N=\log\log N$ for large $N$. Define $A=e^{-a}$, $B=Ne^a$ and
\begin{equation}
 \widetilde q_a(x)=\frac{e^{-Ax}-e^{-Bx}}{(\log N+2a)x}.
 \label{si:padded}
\end{equation}
Frullani's integral shows that this is a probability density. For any $1\le n\le N$, substitute $y=nx$ and set $r=A/n\le e^{-a}$, $s=B/n\ge e^a$. Its second moment is
\begin{align}
 \int_0^\infty\frac{p_n^2}{\widetilde q_a}
 &=(\log N+2a)\int_0^\infty
       \frac{ye^{-2y}}{e^{-ry}-e^{-sy}}\,dy\\
 &=(\log N+2a)\sum_{j=0}^\infty
       \frac1{[2-r+j(s-r)]^2}\\
 &\le(\log N+2a)\left\{\frac1{(2-e^{-a})^2}
              +\frac{\pi^2}{6(e^a-e^{-a})^2}\right\}.
\end{align}
The last bracket tends to $1/4$ uniformly in $n$, proving the matching upper bound. The padded proposal is used only for the unrestricted radius proof; it is not asserted to satisfy a finite candidate-access cap. The bounded envelope proposal supplies the constructive cost bounds.

\paragraph{Finite numerical brackets.}
We partitioned $x$ into 6,001 bins, including the terminal interval to infinity. The bin probabilities
\begin{equation}
 P_{n,j}=e^{-na_j}-e^{-nb_j}
\end{equation}
are exact. Optimizing a simplex prior over a subset of widths produces a dual lower bound on the continuum problem, because coarsening decreases second moments by Cauchy--Schwarz. For the upper bound, a discrete design $q_j$ is lifted using the conditional $\operatorname{Exp}(1)$ law on each bin. Its continuous second moment has the exact expression
\begin{equation}
 V_n^{\mathrm{lift}}=\sum_j\frac{e^{-a_j}-e^{-b_j}}{q_j}
       \frac{n^2}{2n-1}\bigl(e^{-(2n-1)a_j}-e^{-(2n-1)b_j}\bigr).
\end{equation}
Every integer width, including those omitted from optimization, was checked in the upper bound. The bounds below are floating-point evaluations of analytically valid brackets, not interval-arithmetic certificates.
\begin{center}
\begin{tabular}{rrrr}
\toprule
$N$ & Continuum lower & Continuum upper & Bracket gap\\
\midrule
8 & 1.48509132 & 1.48509291 & $1.59\times10^{-6}$\\
32 & 1.83131139 & 1.83131420 & $2.81\times10^{-6}$\\
100 & 2.11729164 & 2.11729489 & $3.25\times10^{-6}$\\
256 & 2.35265023 & 2.35265549 & $5.26\times10^{-6}$\\
1024 & 2.69920131 & 2.69922038 & $1.91\times10^{-5}$\\
\bottomrule
\end{tabular}
\end{center}
Finite-$N$ values should not be approximated by $\log N/4$ without checking boundary effects. In particular, the asymptote alone substantially understates the radius at $N=100$.

\section*{S4. Proof of the two-resource frontier and cost product}
Write $b_N=N^2/(2N-1)$ and
\[
 \R_{N,\kappa}=\inf_{q:\,0<q\le\kappa,\,\int q=1}\max_{n\le N}\int k_n^2/q.
\]
The finite bounds proved in this section are
\begin{equation}
 \max\{\R_N,b_N/\kappa\}\le\R_{N,\kappa}\le\max\{C_N,2N/\kappa\},
 \qquad N\ge2,\ \kappa\ge1.
 \label{si:capped}
\end{equation}
The unrestricted lower bound is immediate. Since $q\le\kappa$,
\begin{equation}
 V_N(q)\ge\kappa^{-1}\int_0^1N^2u^{2N-2}du=b_N/\kappa.
\end{equation}
For the upper bound there are three cases. If $\kappa\ge N/C_N$, use $q_N^{\mathrm{env}}$, for which $\|q\|_\infty=N/C_N$ and $V_n(q)\le C_N$ for every $n$. If $\kappa<2$, uniform sampling gives $\max_nV_n=b_N\le N\le2N/\kappa$. In the remaining case $2\le\kappa<N/C_N$, let
\begin{equation}
 s=\frac{\kappa-1}{N/C_N-1},\qquad
 q=(1-s)+s q_N^{\mathrm{env}}.
\end{equation}
Then $\int q=1$, $0<s<1$, and $\|q\|_\infty=\kappa$. Also
\begin{equation}
 V_n(q)\le\frac{C_N}{s}
 =\frac{N-C_N}{\kappa-1}\le\frac{2N}{\kappa}.
\end{equation}
This proves the finite sandwich. Its uniform order statement follows from $\R_N\sim\log N/4$, $C_N\sim\log N/e$, and $b_N\sim N/2$.

If true percentiles are available, accept each ordinary candidate with probability $q(U)/\kappa$. The acceptance probability is $1/\kappa$, accepted ranks have density $q$, and the conditional truth law is still $f$ because acceptance does not use $Y$. Independent repetition until $T$ accepted candidates has expected candidate count $\kappa T$. Choose $\kappa=\|q\|_\infty$ to avoid an unnecessarily inefficient implementation.

For $V(q)=\max_nV_n(q)$, the exact worst squared error is $V(q)/(4T)$ and the expected cost is $(c_Y+c_X\kappa)T$. Their product is independent of $T$. The lower bounds $V(q)\ge\R_N$ and $\kappa V(q)\ge b_N$ give
\begin{equation}
 (c_Y+c_X\kappa)V(q)\ge c_Y\R_N+c_Xb_N.
\end{equation}
For the envelope, $V(q)\le C_N$ and $\kappa=N/C_N$, yielding the stated upper bound. The constants describe this rejection/importance experiment; methods that reuse several record labels within one path are not covered by its optimality claim. Generation counts are expectations, while a high-probability candidate cap would need additional stopping-time accounting.

\paragraph{Robustness to relative costs.}
Let $\mathcal E_N=\inf_q(c_Y+c_X\|q\|_\infty)\max_nV_n(q)/4$, with the infimum restricted to the centered importance estimator and rejection experiment described in this section. For any nonnegative prices not both zero, divide the envelope upper bound by the optimal lower bound to obtain
\begin{equation}
 \frac{\text{envelope cost-risk product}}{\mathcal E_N}
 \le\frac{c_YC_N+c_XN}{c_Y\R_N+c_Xb_N}
 \le\max\left\{\frac{C_N}{\R_N},\frac{N}{b_N}\right\}.
\end{equation}
The two ratios tend to $4/e$ and $2$. Since the last expression is independent of the prices, the asymptotic bound is uniform even when prices change with $N$. For $c_X=0$, only the first ratio is needed. The factor is asymptotic; the displayed finite ratio supplies the corresponding finite-$N$ bound.

\section*{S5. Simultaneous finite-sample inference}
For a known finite family $G$ and outcome-independent proposal $q$, define
\begin{equation}
 W_n=\esssup k_n/q,\qquad b_n=(W_n+1)/2,\qquad
 x=\log(2|G|/\alpha).
\end{equation}
By Eq.~\ref{si:variance}, $\Var(X_n)\le V_n/4$. Also $|X_n-\E X_n|\le b_n$, since $|X_n|\le W_n/2$ and $|\E X_n|\le1/2$. Bernstein's inequality and a union bound \citep{boucheron2013} give simultaneous radius
\begin{equation}
 r_n=\frac{b_nx}{3T}+
    \sqrt{\frac{V_nx}{2T}+\left(\frac{b_nx}{3T}\right)^2}.
 \label{si:bernstein}
\end{equation}
Intersecting $[\widehat\theta_n-r_n,\widehat\theta_n+r_n]$ with $[0,1]$ preserves coverage. For the envelope, $W_n,V_n\le C_N$; at small desired error the sufficient label count is of order $C_N\varepsilon^{-2}\log(2|G|/\alpha)$.

\paragraph{A sharp local modulus in search width.}
For real $m>n>0$, let $r=m/n$. The densities $k_m,k_n$ cross once at $u_*=(n/m)^{1/(m-n)}$, so
\begin{equation}
 \TV(k_m,k_n)=u_*^n-u_*^m=(r-1)r^{-r/(r-1)}.
\end{equation}
For the logarithmic derivative,
\begin{equation}
 \int_0^1\left|\frac{\partial k_n}{\partial\log n}\right|du
 =\int_0^\infty |1-y|e^{-y}dy=\frac2e.
\end{equation}
Integration along $\log n$ gives $\TV(k_m,k_n)\le |\log(m/n)|/e$. The constant is locally sharp as $r\downarrow1$. The bound on truth means follows because $0\le f\le1$.

\paragraph{Integer grid construction.}
Start with $g_1=1$ and repeatedly set
$g_{j+1}=\min\{N,\lfloor e^\eta g_j\rfloor+1\}$ until reaching $N$. For integer $n$ between consecutive grid points, its left representative satisfies $n\le e^\eta g(n)$. Every nonterminal step grows by a factor greater than $e^\eta$, so the grid has at most $2+\lceil\log N/\eta\rceil$ points. A grid interval $[\ell_g,h_g]$ becomes $[\ell_g-\eta/e,h_g+\eta/e]\cap[0,1]$ for those intermediate targets. No smoothness of $f$ is required.

\paragraph{Post-audit choice.}
Let $\widehat n$ be any measurable function of the complete audit. On the event that all intervals cover, the interval indexed by $\widehat n$ covers its target as well. This protects selection of width within the fixed family. It does not protect selecting a new generator, scorer, or prompt population outside that family. If an entire-curve estimate has uniform error at most $r$, maximizing that estimate incurs reliability regret at most $2r$ within the covered family. This is a consequence of simultaneous validity, not a new post-selection theorem.

\section*{S6. Record paths: correctness, label costs, and coverage}
Give every candidate an independent continuous tie key and rank lexicographically by score and key. For each path, let $I_n$ indicate that draw $n$ is the new maximum, with $I_1=1$. In an iid sequence the rank of each arriving observation among those seen so far is independently uniform on $\{1,\ldots,n\}$. One way to see this is the bijection between permutations and their successive insertion ranks. Consequently the $I_n$ are independent Bernoulli variables with means $1/n$, the classical iid record law \citep{nevzorov2001}.

Store the current winner's label and update it exactly at record times. Every prefix winner is then known after querying $\sum_n I_n$ record-occurrence labels. A label attached to an incumbent is reused without pretending that this creates new independent labels. With unrestricted labels across distinct occurrences, necessity follows by flipping an unqueried record's label. With known identical programs and a deterministic truth oracle, caching can reduce calls further; the occurrence count remains an upper bound.

Across $B$ independent paths, $\E L_B=BH_N$ and $\Var(L_B)\le BH_N$. For $0<\beta<1$, a convenient label-count bound is
\begin{equation}
 \Prb\left\{L_B>BH_N+\sqrt{2BH_N\log(1/\beta)}
                         +\frac23\log(1/\beta)\right\}\le\beta.
\end{equation}
Thus $BH_N$ is an expected cost, not a guaranteed fixed number of labels. Choosing $B$ to fit the displayed upper bound provides an explicit budget-overrun probability; silently stopping early and reusing the fixed-$B$ band would be invalid.

For each width, $Z_b(n)$ is binary with mean $\theta_n$ and is independent across paths. Hoeffding's inequality \citep{hoeffding1963}, followed by a union bound on a fixed grid, proves the record-audit band. Dependence between widths within one path is allowed. The logarithmic-grid extension is Section S5. Setting $\eta=e\varepsilon/2$ and $B\ge2\varepsilon^{-2}\log(2|G|/\alpha)$ makes the sum of sampling and interpolation errors at most $\varepsilon$.

For independently sampled tasks, draw a task once per path and then generate its candidates conditionally iid. The record law holds conditionally on every task, and independent task/path draws make the concentration argument valid for the macro target. Replaying an empirical task pool proves a statement conditional on that pool, not a new-task guarantee.

\section*{S7. Minimax bounds allowing adaptive truth acquisition}
\subsection*{S7.1. Experiment, budgets, and finite statement}
The population percentile law $U\sim\mathrm{Unif}(0,1)$ is known. Candidate draws are iid pairs $(U,Y)$ with $Y\mid U\sim\mathrm{Ber}(f(U))$. The audit observes each inspected rank, can request its truth once, and can retain unqueried candidates for later labeling. Sampling, label selection, and stopping may depend on its random seed and the observed history. The unknown $f$ enters only through truth labels. Auxiliary information supplied to the audit must be independent of $f$; in particular, it does not include population reliability means. Write $K$ for the candidate count and $L$ for the truth-query count. Algorithms stop almost surely and obey $\sup_f\E_fK\le M$ and $\sup_f\E_fL\le T$. There is no assumed parametric, smoothness, or monotonicity structure on $f$.

For $J_N=1+\lfloor\log_{16}N\rfloor$, the following explicit bounds imply Theorem~\ref{th:adaptive}:
\begin{equation}
 \frac1{4096}\min\!\left\{1,\max\!\left(\frac{J_N}{T},\frac{N}{M}\right)\right\}
 \le\mathfrak R_N(M,T)
 \le\min\!\left\{\frac14,\max\!\left(\frac{C_N}{4T},\frac{N}{2M}\right)\right\}.
 \label{si:adaptivefinite}
\end{equation}
The constants are deliberately conservative. The result compares universal orders, not sharp constants for adaptive procedures. Expected-resource constraints are essential to the stated upper construction; a deterministic candidate cap is a different experiment.

\subsection*{S7.2. Disjoint score scales force a logarithmic label cost}
We first give the audit more power: it may choose any $x=-\log u\ge0$ at each query and receive a fresh independent $\mathrm{Ber}(f(e^{-x}))$ response. All unlabeled information is free. Every candidate-based audit can be simulated in this query experiment, so a lower bound here applies to the candidate experiment.

Let $J=J_N$ and choose integer widths $n_j=16^j$, $j=0,\ldots,J-1$. Define disjoint intervals
\begin{equation}
 I_j=[1/(4n_j),4/n_j),\qquad
 a_{ij}=\int_{I_j}n_i e^{-n_i x}\,dx.
\end{equation}
Successive intervals share only an endpoint, and each target places mass
\begin{equation}
 a_{ii}=e^{-1/4}-e^{-4}>\frac34
 \label{si:diagonal}
\end{equation}
on its own interval. For $\omega\in\{-1,+1\}^J$ and $0<\delta\le1/4$, set
\begin{equation}
 f_\omega(e^{-x})=\frac12+\delta\sum_{j=0}^{J-1}\omega_j\one_{I_j}(x).
 \label{si:hypercube}
\end{equation}
Every such law is valid. Since the intervals are disjoint and the total target mass is one,
\begin{equation}
 \omega_i(\theta_{n_i}(f_\omega)-\tfrac12)
 =\delta\left(a_{ii}+\sum_{j\ne i}\omega_i\omega_ja_{ij}\right)
 \ge\delta(2a_{ii}-1)>\frac\delta2.
 \label{si:decodegap}
\end{equation}
Thus the sign of the $i$th target relative to $1/2$ reveals the $i$th bit, regardless of all other bits.

Let $P_\omega^{\mathcal A}$ be the law of the stopped transcript and $L_i$ the number of queries in $I_i$. Flipping only bit $i$ changes the response law only for queries in that interval. The sequential likelihood-ratio chain rule, valid with adaptive actions and stopping \citep{kaufmann2016}, yields
\begin{align}
 \KL(P_\omega^{\mathcal A}\|P_{\omega^{(i)}}^{\mathcal A})
 &=d_\delta\E_\omega L_i,\notag\\
 d_\delta&=2\delta\log\frac{1+2\delta}{1-2\delta}\le16\delta^2.
 \label{si:adaptiveKL}
\end{align}
For $\delta\le1/4$, the last inequality follows from
$\log[(1+z)/(1-z)]\le2z/(1-z)$ at $z=2\delta$.
Action probabilities cancel from the transcript likelihood ratio because the same algorithm is used in both worlds. The identity for unbounded stopping follows by truncation: per-query log ratios are bounded, and expected query counts are finite. Internal randomness and observations whose law is independent of $f$ add no divergence.

Decode $\widehat\omega_i$ by thresholding $\widehat\theta_{n_i}$ at $1/2$, with either fixed convention at equality. Wrong decoding incurs squared error at least $\delta^2/4$. Pairing adjacent hypercube vertices and applying Pinsker's inequality gives
\begin{align}
 \frac1{J2^J}\sum_{\omega,i}\Prb_\omega(\widehat\omega_i\ne\omega_i)
 &\ge\frac12\left(1-\frac1{J2^J}\sum_{\omega,i}
        \TV(P_\omega^{\mathcal A},P_{\omega^{(i)}}^{\mathcal A})\right)\notag\\
 &\ge\frac12\left(1-\sqrt{\frac{8\delta^2T}{J}}\right),
 \label{si:assouad}
\end{align}
because $\sum_i\E_\omega L_i\le T$ for every $\omega$ and Jensen's inequality bounds the average square root. This is a hypercube reduction using stopped-transcript divergences, rather than an assumption that the adaptive observations are iid.

Choose
\begin{equation}
 \delta=\frac18\min\{1,\sqrt{J/T}\}.
\end{equation}
The average decoding error in Eq.~\ref{si:assouad} exceeds $1/4$. The worst-coordinate risk dominates the average coordinate risk under the uniform hypercube prior, giving
\begin{equation}
 \sup_f\max_{n\le N}\E_f(\widehat\theta_n-\theta_n)^2
 \ge\frac1{1024}\min\{1,J_N/T\}.
 \label{si:labellower}
\end{equation}
This bound permits outcome-adaptive acquisition and arbitrary estimators. It does not identify the sharp adaptive constant $1/4$ appearing in the restricted importance-design radius.

\subsection*{S7.3. Rare truth requires candidate access even with known percentiles}
Use the same uniform score law in two worlds and set
\begin{equation}
 f_0(u)=\tfrac12,\qquad
 f_1(u)=\tfrac12+\delta\one_{[1-1/N,1]}(u),\qquad0<\delta\le\tfrac14.
\end{equation}
Their width-$N$ targets differ by
\begin{equation}
 \Delta=\delta\{1-(1-1/N)^N\}\ge(1-e^{-1})\delta>\delta/2.
\end{equation}
Grant the audit the truth of every inspected candidate for free. For one fully observed pair,
\begin{equation}
 \KL(P_0\|P_1)
 =\frac1N\KL(\mathrm{Ber}(1/2)\|\mathrm{Ber}(1/2+\delta))
 =-\frac1{2N}\log(1-4\delta^2)\le\frac{4\delta^2}{N}.
\end{equation}
The stopped-transcript divergence is at most $4\delta^2\E_0K/N\le4\delta^2M/N$. One may equivalently augment the actual transcript with the unused candidate truths and apply data processing; the algorithm's stopping rule remains a stopping time for the augmented filtration. With $\delta=\tfrac18\min\{1,\sqrt{N/M}\}$, this divergence is at most $1/16$.

Testing the two worlds by choosing the closer target gives the standard two-point squared-error inequality \citep{tsybakov2009},
\begin{equation}
 \max_{a\in\{0,1\}}\E_a(\widehat\theta_N-\theta_N(f_a))^2
 \ge\frac{\Delta^2}{8}\{1-\TV(P_0^{\mathcal A},P_1^{\mathcal A})\}
 \ge\frac1{4096}\min\{1,N/M\}.
 \label{si:candidatelower}
\end{equation}
Here Pinsker gives total variation at most $\sqrt{1/32}<1/2$, and $\Delta>\delta/2$. The population score distribution is identical and known in both worlds: learning the score law cannot resolve the hidden truth difference. Only ordinary iid candidate access is allowed; an oracle that directly produces any desired rank would remove this resource constraint.

\subsection*{S7.4. Matching upper bound and cost consequence}
The maximum of Eqs.~\ref{si:labellower} and \ref{si:candidatelower} proves the lower half of Eq.~\ref{si:adaptivefinite}. For its upper half, take $\kappa=M/T$ and use the density-capped construction of Section S4. Rejection sampling until $T$ accepted labels has expected candidate count at most $\kappa T=M$, and Eq.~\ref{si:variance} gives risk at most
\begin{equation}
 \frac1{4T}\max\{C_N,2N/\kappa\}
 =\max\{C_N/(4T),N/(2M)\}.
\end{equation}
If this exceeds $1/4$, the constant estimator $1/2$ uses no observations and has risk at most $1/4$. Since $J_N\asymp1+\log N$, $C_N\asymp1+\log N$, and the maximum and sum of two nonnegative numbers differ by at most a factor two, Eq.~\ref{si:adaptivefinite} proves the uniform order in Theorem~\ref{th:adaptive}.

For a sufficiently small fixed target mean squared error $\varepsilon^2$, the separate lower bounds force $T\gtrsim(1+\log N)/\varepsilon^2$ and $M\gtrsim N/\varepsilon^2$. The envelope uses $T=\lceil C_N/(4\varepsilon^2)\rceil$ and $M=(N/C_N)T$, meeting both orders simultaneously. Nonnegative costs $c_X,c_Y$, not both zero, therefore give the main-text cost law uniformly over their relative sizes. This establishes an order statement for all admissible adaptive procedures. The more precise asymptotic factor-two guarantee in Section S4 concerns only its explicitly specified rejection/importance class.

\subsection*{S7.5. Necessary resources for simultaneous high-probability accuracy}
For $0<\varepsilon\le1/16$ and $0<\alpha<1/4$, suppose an audit estimates every width within $\varepsilon$ simultaneously with probability at least $1-\alpha$, for every admissible law. In Eq.~\ref{si:hypercube}, choose $\delta=4\varepsilon$. By Eq.~\ref{si:decodegap}, each bit is then decoded correctly on the simultaneous event. For every $\omega$ and $i$, a test between $\omega$ and $\omega^{(i)}$ has sum of errors at most $2\alpha$. The two-point testing inequality and Eq.~\ref{si:adaptiveKL} imply
\begin{equation}
 16\delta^2\E_\omega L_i\ge\log\frac1{4\alpha}.
\end{equation}
Summing over $i$ yields a necessary worst-case expected label budget
\begin{equation}
 T\ge\frac{J_N}{256\varepsilon^2}\log\frac1{4\alpha}.
 \label{si:confidence-label-lower}
\end{equation}
Using $\delta=4\varepsilon$ in Section S7.3 gives $\Delta>2\varepsilon$ and divergence at most $64M\varepsilon^2/N$, hence
\begin{equation}
 M\ge\frac{N}{64\varepsilon^2}\log\frac1{4\alpha}.
\end{equation}
These bounds show that both resource orders also enter simultaneous certification. They leave a gap in the extra logarithmic factor used by the grid-based upper construction. They do not establish its necessity.


\section*{S8. Finite pools, ties, and reproducibility}
\paragraph{Exact target probabilities.}
Within task $t$, a score-tie group of size $h$ among $M$ candidates occupies percentile interval $[a,b]$, where $b-a=h/M$. Each candidate $i$ in the group has winner probability
\begin{equation}
 p_{n,t,i}=\frac{b^n-a^n}{h}.
\end{equation}
For $K$ tasks, the macro target kernel is $P_n(t,i)=p_{n,t,i}/K$. The 5\% tail law assigns each tied candidate probability
$[b-\max\{a,.95\}]_+/(.05hK)$, so partial overlap with the cutoff is handled exactly. The envelope is
\begin{equation}
 q^{\mathrm{env}}(t,i)=\frac{\max_{n\le100}P_n(t,i)}{
                           \sum_{t',i'}\max_{n\le100}P_n(t',i')}.
\end{equation}
The finite normalizer can be smaller than the unrestricted $C_N$: taking a maximum after integrating within a score-tie interval cannot exceed integrating the pointwise maximum. The variance radius also contracts under coarsening by Cauchy--Schwarz. No arbitrary truth law was inserted inside a tie group to claim an exact fixed-pool interval.

\paragraph{Protocol.}
The acquisition specification fixes all integer widths 1--100, 500 draws, 2,000 replays, 95\% simultaneous coverage, seed 20260904, and five score-based proposals. The shared discovery/held-out partition is fixed by seed 260821. All scores in the 8,200-candidate held-out pool per model are available when constructing the proposal. The top-tail and winner-100 comparators each mix their focused distribution with 20\% uniform coverage. These are retrospective analyses of public candidate collections; the computational specification is not a public preregistration.

\paragraph{Outcome-independent design and uncertainty.}
The optimization function accepts only the kernel matrix, never labels. Acquisition indices are iid from the resulting fixed proposal. Their randomness supports conditional finite-pool coverage even when the same program is sampled twice. It does not turn repeated candidate observations into independent additional benchmark tasks. We report both 500 draws and expected unique identities,
\begin{equation}
 \E[\text{unique labels}]=\sum_{t,i}\{1-[1-q(t,i)]^{500}\}.
\end{equation}
The primary quantile summarizes acquisition randomness conditional on the task pool. We make no confidence claim about generalization to a larger task population. With 2,000 replays, Monte Carlo uncertainty in the nearly identical envelope and variance-optimal error quantiles precludes interpreting their small differences as a substantive ranking. The maximum absolute estimated mean bias was below 0.0017 across designs; exact unbiasedness follows analytically.

\paragraph{What coverage means.}
All ten real-data method/model cells had full simultaneous and selected-width coverage in the 2,000 replays. This is compatible with deliberately conservative Bernstein intervals. The guarantee is at least 95\%; this conservative construction does not target exact 95\% calibration. Synthetic coverage was 1.000, 1.000, 0.999, and 1.000 for constant, rare-tail-failure, smooth, and oscillatory laws respectively.

\begin{table}[ht]
\centering
\caption{\textbf{Auditing all widths 1--100 with 500 label draws.} $Q_{.95}$ is the 95th percentile, over acquisition replays, of the maximum absolute error over widths. Width is the median of the maximum clipped simultaneous interval width. Coverage was 100\% in these 2,000 replays for each listed design; the analytic guarantee is at least 95\%, and the bands are conservative. Distinct labels are expected unique program identities, not additional independent tasks.}
\label{tab:empirical}
\begin{tabular}{llrrr}
\toprule
Model & Design & $Q_{.95}$ maximum error & Maximum band width & Distinct labels\\
\midrule
8B & Uniform & 0.1435 & 0.8307 & 485.1\\
   & Top 5\% + uniform & 0.0766 & 0.4900 & 453.5\\
   & Winner-100 + uniform & 0.0780 & 0.4050 & 429.8\\
   & Envelope & 0.0596 & 0.2415 & 449.0\\
   & Minimax variance & 0.0595 & 0.2405 & 453.0\\
\midrule
70B & Uniform & 0.0789 & 0.6327 & 485.1\\
    & Top 5\% + uniform & 0.0574 & 0.4146 & 471.1\\
    & Winner-100 + uniform & 0.0555 & 0.3334 & 467.4\\
    & Envelope & 0.0477 & 0.2160 & 469.4\\
    & Minimax variance & 0.0477 & 0.2383 & 472.9\\
\bottomrule
\end{tabular}
\end{table}

\paragraph{Record replay.}
Each of 100 paths first samples a held-out task uniformly, then samples 100 candidate indices with replacement. An independent key breaks score ties on each occurrence. All prefix holders are reconstructed, and labels are requested only at record arrivals. The label count and unique identities are both logged. The simple simultaneous band for all 100 widths has unclipped width 0.407285 at this path count. The experiment measures reuse within an existing scored pool and does not constitute fresh generation or measured human labeling cost.

\paragraph{Computational provenance.}
The source package records the processed-input hashes, source-release identifiers, random seeds, and Python dependency versions. Candidate-level code results are recomputed from the numeric score--truth table. The mathematical-reasoning workflow uses processed search spectra and problem summaries; rebuilding scores from the large raw generations is a separate documented workflow. The lightweight scripts make no model calls. The acquisition simulation uses Python 3.12.13, NumPy 2.3.5, SciPy 1.17.0, and Matplotlib 3.10.8.

\begin{figure}[ht]
\centering\includegraphics[page=4,width=0.94\textwidth]{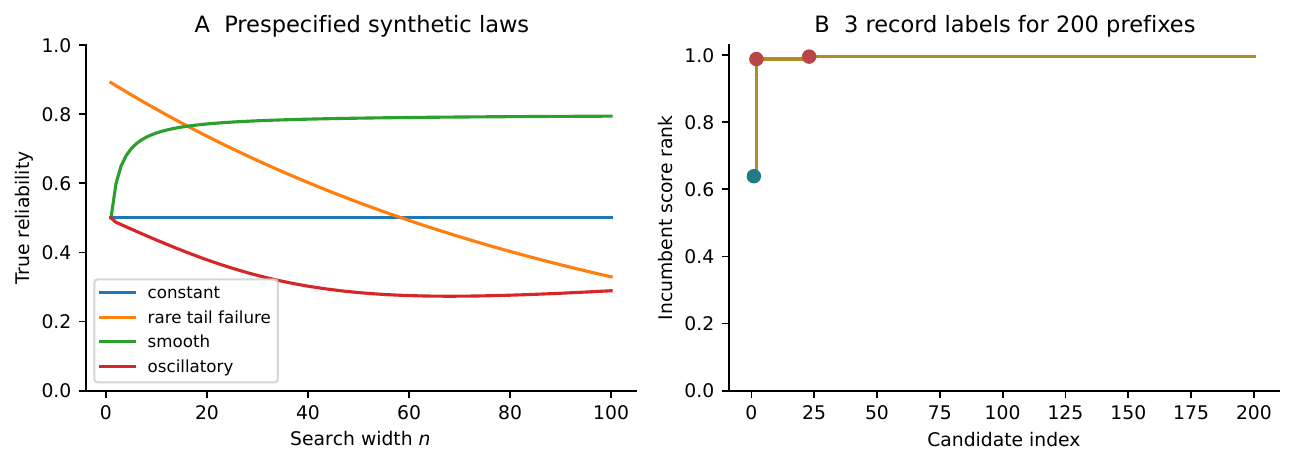}
\caption{\textbf{Implementation checks for distinct truth mechanisms.} (A) The synthetic surfaces produce different complete reliability curves, including a rare high-score failure region. (B) A simulated path illustrates that incumbent labels can be reused across many prefixes. Colors distinguish true and false record candidates. This is a generated stochastic example, not an empirical model trace.}
\end{figure}

\section*{S9. Two-domain evidence, compatibility, and fixed-target acquisition}
\subsection*{S9.1. Mathematical-reasoning data and uncertainty}
The Monkey Business release \citep{brown2024}, revision \texttt{a9f8f73bcd6948a57ed922cba4e48062ef95f553}, supplies 127 GSM8K problems per model, each with 10,000 generated solutions. Both release files are included in full in the analysis population. Their SHA-256 hashes and reconstruction URLs are embedded in \texttt{upstream/code/analyze\_search\_spectrum.py}. Within each model--problem cell, 2,000 solutions define reference answer frequencies and a disjoint 8,000 solutions define the empirical deployment population. Final answers are normalized by removing commas and whitespace. Unparseable answers receive a score below parsed answers. The split seed is 20260817 with fixed model and problem offsets.

For score group $\ell$, let $F_{\ell-},F_\ell$ denote cumulative masses immediately below and through that group and let $t_\ell$ denote its mean truth. The exact with-replacement selected truth is
\begin{equation}
 \theta_n^{\mathrm{pool}}=\sum_\ell t_\ell(F_\ell^n-F_{\ell-}^n).
\end{equation}
We evaluate widths $1,2,4,\ldots,4096$. Figure~\ref{fig:domains} uses the stored numeric search spectra, with 20,000 problem-bootstrap replicates (seed 73041 plus model offset). Ten distinct reference/deployment splits test stability of the harm counts. All 127 problems enter search and harm summaries; AUC is computed only for the 125 mixed-label 8B and 107 mixed-label 70B populations. The two and 20 single-class populations remain in the other summaries.

\begin{center}
\begin{tabular}{lrr}
\toprule
GSM8K quantity & 8B & 70B\\
\midrule
Mean truth, width 1 & 0.76835 & 0.93053\\
Mean truth, width 4096 & 0.88189 & 0.96850\\
Macro within-problem AUC & 0.90698 & 0.96166\\
Problems harmed by more than 1 pp & 14 & 3\\
Worst change & $-0.479625$ & $-0.359750$\\
Harm count across ten splits & 13--15 & 3 in every split\\
\bottomrule
\end{tabular}
\end{center}

\subsection*{S9.2. Code-generation data}
CodeRM \citep{ma2025} is pinned to commit \texttt{aa4946e9245ed41e24d60ad29e965132b5b84fe6}. The analysis uses 100 candidate programs for every one of 164 HumanEval+ tasks per model. Scores are the fractions of 100 generated unit tests passed. The separate HumanEval+ \texttt{plus\_status} is the binary truth label. Thus the source records 32,800 programs and 3.28 million candidate--unit-test executions. The package contains derived numeric scores and labels; raw programs and execution logs are obtained through the checksum-verifying reconstruction script.

Task-specific tie-averaged kernels yield search curves and AUCs. The bootstrap samples tasks, with 20,000 replicates and seed 92771 plus model offset. All tasks contain both truth classes in each model. Means, AUCs, harm counts, and worst losses are recomputed from the candidate-level numeric table. Candidate and unit-test counts are descriptive sample sizes, not counts of independent benchmark tasks.

\begin{center}
\begin{tabular}{lrr}
\toprule
CodeRM quantity & 8B & 70B\\
\midrule
Mean truth, width 1 & 0.53579 & 0.73744\\
Mean truth, width 100 & 0.71508 & 0.78844\\
Macro within-task AUC & 0.931 & 0.881\\
Tasks harmed by more than 1 pp & 10 & 4\\
Worst change & $-0.722$ & $-0.600$\\
\bottomrule
\end{tabular}
\end{center}

\subsection*{S9.3. Bounded compatibility witnesses}
Within each task, distribute each score-tie group's mean truth uniformly over its occupied percentile intervals. The task-averaged step function $\bar f$ then reproduces every macro search mean. To measure what a prefix alone reveals, use $J=1000$ equal bins and candidate reliability values $g_j\in[0,1]$. Define
\begin{equation}
 w_{n,j}=(j/J)^n-((j-1)/J)^n.
\end{equation}
Minimize and maximize $w_N^\top g$ subject to $w_n^\top g=\widehat\theta_n$ for $n=1,\ldots,m$. Each optimizer is a valid bounded continuum step function. Its separation therefore witnesses a lower bound on the unrestricted continuum diameter. A QR basis of the audit row space controls numerical conditioning without changing the equality constraints. At $m=8$, maximum raw moment residuals are below $1.3\times10^{-14}$; changing the grid from 200 to 1000 bins moves each endpoint by less than 0.007.

Task uncertainty is assessed separately with 20,000 multinomial task-bootstrap replicates. For each prefix, a 95th-percentile maximum studentized deviation supplies simultaneous audit bands. Substituting these bands for exact equalities yields the bootstrap-feasible ranges. These bootstrap bands are conditional empirical uncertainty summaries, distinct from the distribution-free acquisition bands in Section S5.

\begin{table}[ht]
\centering
\caption{\textbf{CodeRM best-of-100 compatibility from aggregate audit means.} The endpoints are attained by bounded 1000-bin laws. They are constructive witnesses, not exact endpoints of the unrestricted continuum problem.}
\begin{tabular}{lrrr}
\toprule
Model & Audited prefix $m$ & Exact plug-in range & Bootstrap-feasible range\\
\midrule
8B & 8 & [0.186, 0.992] & [0.003, 1.000]\\
8B & 20 & [0.690, 0.737] & [0.102, 0.999]\\
70B & 8 & [0.305, 0.996] & [0.016, 1.000]\\
70B & 20 & [0.769, 0.806] & [0.197, 1.000]\\
\bottomrule
\end{tabular}
\end{table}

\begin{figure}[ht]
\centering\includegraphics[page=5,width=0.93\textwidth]{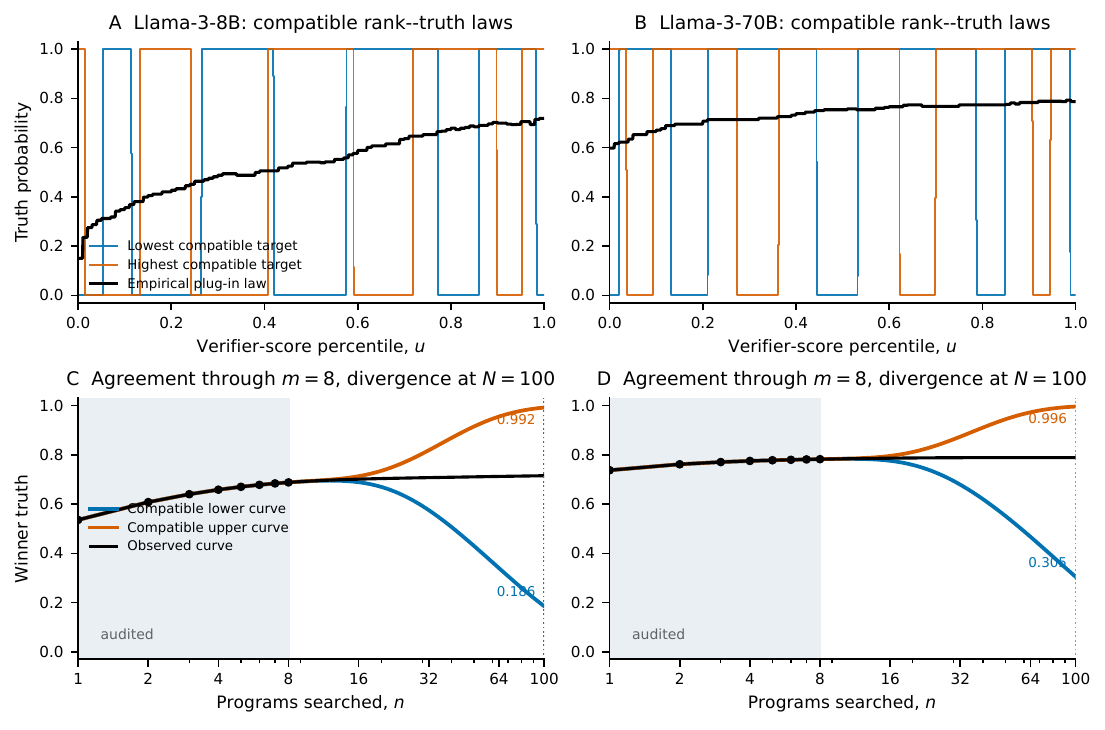}
\caption{\textbf{Different deployment truths can match the same empirical audit prefix.} The bounded rank--truth witnesses match every CodeRM mean at widths 1--8. Their induced search curves separate at larger widths. The constructions demonstrate compatibility with the retained evidence; they do not estimate how frequently these extreme reliability laws occur.}
\end{figure}

\subsection*{S9.4. Discovery-selected fixed-target audit}
A seeded permutation (260821) divides shared CodeRM task identifiers into 82 discovery and 82 held-out tasks. On discovery tasks, the fixed candidate family comprises top 5\%, 10\%, and 20\% score-tail audits. The choice minimizes mean 95th-percentile absolute best-of-100 error across the two models at 500 labels and 5000 replays. The 5\% rule is selected and then frozen before held-out targets and errors are computed.

If $P_{100}(t,i)$ is the fixed-target kernel and $q(t,i)$ the chosen tail proposal, the estimator is $T^{-1}\sum P_{100}(t_i,i_i)Y_i/q(t_i,i_i)$ on supported candidates. Its omitted target mass is bounded by $0.95^{100}=0.0059205$. The small observed bias is reported alongside this distribution-free bound. Independent replay streams use seed 262147 plus fixed offsets.

\begin{table}[ht]
\centering
\caption{\textbf{A focused audit for a prespecified best-of-100 target.} These errors concern one target and an uncentered, truncated-support estimator. They are not the complete-curve errors in Fig.~\ref{fig:empirical}.}
\begin{tabular}{lrrrr}
\toprule
Model & Held-out truth & Uniform $Q_{.95}$ error & Frozen-tail $Q_{.95}$ error & Absolute tail bias\\
\midrule
8B & 0.698 & 0.180 & 0.049 & $7.58\times10^{-5}$\\
70B & 0.791 & 0.076 & 0.035 & $7.28\times10^{-5}$\\
\bottomrule
\end{tabular}
\end{table}

The experiment assesses transfer of a discovery-selected allocation rule within one public dataset. The complete-curve experiment instead uses score-only proposals and a centered estimator with support over every width. The experiments share a task split but answer different questions. Neither is a prospective intervention measuring realized acquisition costs.


\section*{S10. Structural ambiguity under shape constraints}
The structural gap can persist for monotone Lipschitz reliability laws. Let $\mathcal F_{\uparrow,L}$ contain nondecreasing $L$-Lipschitz maps from $[0,1]$ to $[0,1]$, and let $\Delta^{\uparrow,L}_{m,N}$ be the largest separation at width $N>m$ between two such laws agreeing at widths $1,\ldots,m$.

\paragraph{Constructive finite bounds.}
Put
\begin{equation}
 A_{m,L}=\min\{1/(m+1),L/[m(m+1)]\},\qquad
 D_{m,N}=\prod_{j=1}^m\frac{N-j}{N+j}.
\end{equation}
Let $d=\lfloor(m-1)/2\rfloor$ and let $\xi_{d+1}$ be the largest shifted Legendre zero of degree $d+1$. Then
\begin{equation}
 A_{m,L}D_{m,N}\le\Delta^{\uparrow,L}_{m,N}
 \le\min\left\{B_{m,N},L\left(1-\xi_{d+1}+\frac1{N+1}\right)\right\}.
 \label{si:shapebounds}
\end{equation}

For the lower bound, use the shifted Legendre polynomial $L_m(u)=P_m(2u-1)$ and define
\begin{align}
 f(u)&=A_{m,L}[L_m(0)]_++A_{m,L}\int_0^u[L_m'(t)]_+dt,\notag\\
 g(u)&=A_{m,L}[-L_m(0)]_++A_{m,L}\int_0^u[-L_m'(t)]_+dt.
\end{align}
Then $f-g=A_{m,L}L_m$. Both functions are nonnegative and nondecreasing. The bound $\|L_m'\|_\infty=m(m+1)$ makes them $L$-Lipschitz. Since $|L_m|\le1$ and it has $m$ monotonicity intervals, its variation is at most $2m$; each displayed function is at most $A_{m,L}(m+1)\le1$. Orthogonality gives agreement for all $n\le m$. Rodrigues' identity and $m$ integrations by parts give
\begin{align}
 \int_0^1Nu^{N-1}L_m(u)du
 &=\frac{N[(N-1)!]^2}{(N-1-m)!(N+m)!}
 =D_{m,N},
\end{align}
proving the lower bound. The functions are continuously differentiable with Lipschitz truth profiles; they do not require a discontinuous Chebyshev-sign law.

For the upper bound, $h=f-g$ is $L$-Lipschitz: the increments of each monotone law over an interval of length $a$ lie in $[0,La]$. It also annihilates all degree-$(m-1)$ polynomials. Among $p\in\mathcal P_d$, the minimum of
\begin{equation}
 \frac{\int(1-u)p(u)^2du}{\int p(u)^2du}
\end{equation}
is $1-\xi_{d+1}$, by the multiplication-operator representation of Gauss--Legendre nodes \citep{szego1939}. Normalize an attaining square to $q=p^2/\int p^2$. It has degree at most $m-1$, so $\int qh=0$ and
\begin{equation}
 |h(1)|\le\int q(u)|h(1)-h(u)|du\le L(1-\xi_{d+1}).
\end{equation}
Since the winner rank has expected distance $1/(N+1)$ from one,
$|\int k_Nh|\le L(1-\xi_{d+1}+1/(N+1))$.
The unrestricted diameter supplies the other upper bound in Eq.~\ref{si:shapebounds}.

For fixed $L>0$ and $m^2/N\to\tau\in(0,\infty)$, these bounds give
\begin{equation}
 e^{-\tau}\le\liminf\frac{m^2\Delta^{\uparrow,L}_{m,N}}L
 \le\limsup\frac{m^2\Delta^{\uparrow,L}_{m,N}}L\le j_{0,1}^2+\tau,
\end{equation}
where $j_{0,1}$ is the first positive zero of $J_0$. The lower asymptote follows from $D_{m,N}\to e^{-\tau}$; the upper uses the classical extreme Legendre-zero asymptote \citep{szego1939}. Shape constraints therefore reduce ambiguity to order $L/m^2$ in this regime while preserving the degree-squared endpoint scale.

\paragraph{Evidence-specific identification.}
A monotone law can be represented as a distribution function of a nonnegative measure of total mass at most one. With $b_n(t)=1-t^n$, its exact identified interval given a feasible audit vector $y$ is the range of $\int b_Nd\mu$ subject to $\mu\ge0$, $\mu([0,1])\le1$, and $\int b_nd\mu=y_n$ for $n\le m$. For the Lipschitz class, write $f(u)=c+\int_0^u h(t)dt$, $c\ge0$, $0\le h\le L$, $c+\int h\le1$. The interval is the range of
\begin{equation}
 c+\int_0^1b_Nh,
 \quad c+\int_0^1b_nh=y_n\quad(n\le m).
\end{equation}
These compact convex programs distinguish a realized audit vector from the largest diameter over all feasible vectors. Simultaneous audit bands can replace the equalities to propagate uncertainty. The source package contains the scalar-threshold capped-tail solver and its finite numerical checks.




\begin{thebibliography}{19}
\providecommand{\natexlab}[1]{#1}
\providecommand{\url}[1]{\texttt{#1}}
\expandafter\ifx\csname urlstyle\endcsname\relax
  \providecommand{\doi}[1]{doi: #1}\else
  \providecommand{\doi}{doi: \begingroup \urlstyle{rm}\Url}\fi

\bibitem[Bojanov et~al.(1986)Bojanov, Braess, and Dyn]{bojanov1986}
Borislav~D. Bojanov, Dietrich Braess, and Nira Dyn.
\newblock Generalized gaussian quadrature formulas.
\newblock \emph{Journal of Approximation Theory}, 48\penalty0 (4):\penalty0
  335--353, 1986.
\newblock \doi{10.1016/0021-9045(86)90008-0}.

\bibitem[Boucheron et~al.(2013)Boucheron, Lugosi, and Massart]{boucheron2013}
St{\'e}phane Boucheron, G{\'a}bor Lugosi, and Pascal Massart.
\newblock \emph{Concentration Inequalities: A Nonasymptotic Theory of
  Independence}.
\newblock Oxford University Press, 2013.
\newblock \doi{10.1093/acprof:oso/9780199535255.001.0001}.

\bibitem[Brown et~al.(2024)Brown, Juravsky, Ehrlich, Clark, Le, R{\'e}, and
  Mirhoseini]{brown2024}
Bradley Brown, Jordan Juravsky, Ryan Ehrlich, Ronald Clark, Quoc~V. Le,
  Christopher R{\'e}, and Azalia Mirhoseini.
\newblock Large language monkeys: Scaling inference compute with repeated
  sampling.
\newblock arXiv:2407.21787, 2024.

\bibitem[Chen et~al.(2025)Chen, Pacchiano, and Paschalidis]{chen2025}
Yilei Chen, Aldo Pacchiano, and Ioannis Paschalidis.
\newblock Multiple-policy evaluation via density estimation.
\newblock In \emph{Proceedings of the 42nd International Conference on Machine
  Learning}, volume 267 of \emph{Proceedings of Machine Learning Research},
  pages 8794--8820, 2025.
\newblock URL \url{https://proceedings.mlr.press/v267/chen25aw.html}.

\bibitem[Donoho(1994)]{donoho1994}
David~L. Donoho.
\newblock Statistical estimation and optimal recovery.
\newblock \emph{The Annals of Statistics}, 22\penalty0 (1):\penalty0 238--270,
  1994.
\newblock \doi{10.1214/aos/1176325367}.

\bibitem[Dorner et~al.(2026)Dorner, Chen, Cruz, and Yang]{dorner2026}
Florian~E. Dorner, Yatong Chen, Andr{\'e}~F. Cruz, and Fanny Yang.
\newblock {ROC-n-Reroll}: How verifier imperfection affects test-time scaling.
\newblock In \emph{International Conference on Learning Representations}, 2026.
\newblock URL \url{https://arxiv.org/pdf/2507.12399}.
\newblock arXiv:2507.12399v3, revised 17 August 2026.

\bibitem[Fluri et~al.(2025)Fluri, Lang, Abate, Forr{\'e}, Krueger, and
  Skalse]{fluri2025}
Lukas Fluri, Leon Lang, Alessandro Abate, Patrick Forr{\'e}, David Krueger, and
  Joar Max~Viktor Skalse.
\newblock The perils of optimizing learned reward functions: Low training error
  does not guarantee low regret.
\newblock In \emph{Proceedings of the 42nd International Conference on Machine
  Learning}, volume 267 of \emph{Proceedings of Machine Learning Research},
  pages 17306--17377, 2025.
\newblock URL \url{https://proceedings.mlr.press/v267/fluri25a.html}.

\bibitem[Hoeffding(1963)]{hoeffding1963}
Wassily Hoeffding.
\newblock Probability inequalities for sums of bounded random variables.
\newblock \emph{Journal of the American Statistical Association}, 58\penalty0
  (301):\penalty0 13--30, 1963.
\newblock \doi{10.1080/01621459.1963.10500830}.

\bibitem[Huang et~al.(2025)Huang, Block, Liu, Jiang, Krishnamurthy, and
  Foster]{huang2025}
Audrey Huang, Adam Block, Qinghua Liu, Nan Jiang, Akshay Krishnamurthy, and
  Dylan~J. Foster.
\newblock Is {Best-of-N} the best of them? coverage, scaling, and optimality in
  inference-time alignment.
\newblock In \emph{Proceedings of the 42nd International Conference on Machine
  Learning}, volume 267 of \emph{Proceedings of Machine Learning Research},
  pages 25075--25126, 2025.
\newblock URL \url{https://proceedings.mlr.press/v267/huang25c.html}.

\bibitem[Kaufmann et~al.(2016)Kaufmann, Capp{\'e}, and Garivier]{kaufmann2016}
Emilie Kaufmann, Olivier Capp{\'e}, and Aur{\'e}lien Garivier.
\newblock On the complexity of best-arm identification in multi-armed bandit
  models.
\newblock \emph{Journal of Machine Learning Research}, 17\penalty0
  (1):\penalty0 1--42, 2016.
\newblock URL \url{https://jmlr.org/papers/v17/kaufman16a.html}.

\bibitem[Ma et~al.(2025)Ma, Zhang, Zhang, Yu, Luo, and Tang]{ma2025}
Zeyao Ma, Xiaokang Zhang, Jing Zhang, Jifan Yu, Sijia Luo, and Jie Tang.
\newblock Dynamic scaling of unit tests for code reward modeling.
\newblock In \emph{Proceedings of the 63rd Annual Meeting of the Association
  for Computational Linguistics (Volume 1: Long Papers)}, pages 6917--6935,
  2025.
\newblock \doi{10.18653/v1/2025.acl-long.343}.
\newblock URL \url{https://aclanthology.org/2025.acl-long.343/}.

\bibitem[Nakibo{\u g}lu(2019)]{nakiboglu2019}
Bar{\i}{\c s} Nakibo{\u g}lu.
\newblock The r{\'e}nyi capacity and center.
\newblock \emph{IEEE Transactions on Information Theory}, 65\penalty0
  (2):\penalty0 841--860, 2019.
\newblock \doi{10.1109/TIT.2018.2861002}.
\newblock URL \url{https://arxiv.org/abs/1608.02424}.

\bibitem[Nevzorov(2001)]{nevzorov2001}
Valery~B. Nevzorov.
\newblock \emph{Records: Mathematical Theory}, volume 194 of \emph{Translations
  of Mathematical Monographs}.
\newblock American Mathematical Society, 2001.
\newblock ISBN 978-0-8218-1945-6.
\newblock URL \url{https://bookstore.ams.org/MMONO/194}.

\bibitem[Newman and Rivlin(1976)]{newman1976}
Donald~J. Newman and Theodore~J. Rivlin.
\newblock Approximation of monomials by lower degree polynomials.
\newblock \emph{Aequationes Mathematicae}, 14:\penalty0 451--455, 1976.
\newblock \doi{10.1007/BF01835995}.

\bibitem[Russo and Pacchiano(2025)]{russo2025}
Alessio Russo and Aldo Pacchiano.
\newblock Adaptive exploration for multi-reward multi-policy evaluation.
\newblock In \emph{Proceedings of the 42nd International Conference on Machine
  Learning}, volume 267 of \emph{Proceedings of Machine Learning Research},
  pages 52382--52421, 2025.
\newblock URL \url{https://arxiv.org/abs/2502.02516}.

\bibitem[Shtarkov(1987)]{shtarkov1987}
Yuri~M. Shtarkov.
\newblock Universal sequential coding of single messages.
\newblock \emph{Problems of Information Transmission}, 23\penalty0
  (3):\penalty0 175--186, 1987.
\newblock URL \url{https://www.mathnet.ru/eng/ppi811}.

\bibitem[Szeg{\H{o}}(1975)]{szego1939}
G{\'a}bor Szeg{\H{o}}.
\newblock \emph{Orthogonal Polynomials}, volume~23 of \emph{American
  Mathematical Society Colloquium Publications}.
\newblock American Mathematical Society, Providence, RI, 4 edition, 1975.
\newblock \doi{10.1090/coll/023}.

\bibitem[Tsybakov(2009)]{tsybakov2009}
Alexandre~B. Tsybakov.
\newblock \emph{Introduction to Nonparametric Estimation}.
\newblock Springer, 2009.
\newblock \doi{10.1007/b13794}.

\bibitem[Whittaker and Watson(1927)]{whittaker1927}
E.~T. Whittaker and G.~N. Watson.
\newblock \emph{A Course of Modern Analysis}.
\newblock Cambridge University Press, Cambridge, UK, 4 edition, 1927.

\end{thebibliography}
\end{document}